%% file: template.tex
\documentclass[conference]{IEEEtran}
\IEEEoverridecommandlockouts
\usepackage{cite}
\usepackage{amsmath,amssymb,amsfonts}
\usepackage{algorithmic}
\usepackage{latexsym}
\usepackage{textcomp}
\usepackage{booktabs}
\usepackage{graphicx} 
\usepackage{xcolor}
\def\BibTeX{{\rm B\kern-.05em{\sc i\kern-.025em b}\kern-.08em
    T\kern-.1667em\lower.7ex\hbox{E}\kern-.125emX}}

\usepackage{fancyhdr}
\begin{document}

\title{Context-Adaptive Synthesis and Compression for Enhanced Retrieval-Augmented Generation in Complex Domains}

\author{Peiran Zhou, Junnan Zhu, Yichen Shen, Ruoxi Yu \\
Kunming Medical University}

\maketitle
\thispagestyle{fancy} 

\input{main}

\bibliographystyle{IEEEtran}
\bibliography{references}
\end{document}

%% file: main.tex
\begin{abstract}
Large Language Models (LLMs) excel in language tasks but are prone to hallucinations and outdated knowledge. Retrieval-Augmented Generation (RAG) mitigates these by grounding LLMs in external knowledge. However, in complex domains involving multiple, lengthy, or conflicting documents, traditional RAG suffers from information overload and inefficient synthesis, leading to inaccurate and untrustworthy answers. To address this, we propose CASC (Context-Adaptive Synthesis and Compression), a novel framework that intelligently processes retrieved contexts. CASC introduces a Context Analyzer \& Synthesizer (CAS) module, powered by a fine-tuned smaller LLM, which performs key information extraction, cross-document consistency checking and conflict resolution, and question-oriented structured synthesis. This process transforms raw, scattered information into a highly condensed, structured, and semantically rich context, significantly reducing the token count and cognitive load for the final Reader LLM. We evaluate CASC on SciDocs-QA, a new challenging multi-document question answering dataset designed for complex scientific domains with inherent redundancies and conflicts. Our extensive experiments demonstrate that CASC consistently outperforms strong baselines.
\end{abstract}

\section{Introduction}

Large Language Models (LLMs) have demonstrated remarkable capabilities in understanding, generating, and processing human language across a wide array of tasks \cite{yifan2023a}. However, despite their impressive fluency and reasoning abilities, LLMs inherently suffer from several critical limitations, including the propensity for "hallucinations" (generating factually incorrect or nonsensical information) and a knowledge cut-off date, rendering them incapable of accessing the most current information \cite{yao2024a}. Retrieval-Augmented Generation (RAG) has emerged as a powerful paradigm to mitigate these issues by grounding LLM responses in external, up-to-date knowledge bases \cite{patrick2020retrie}. By dynamically retrieving relevant information and injecting it into the LLM's context, RAG systems significantly enhance factual accuracy, reduce hallucination, and improve the trustworthiness of generated content. This approach is particularly crucial for applications in complex domains such as scientific research, legal analysis, and medical diagnostics, where precision and verifiable facts are paramount.

Despite the promise of RAG, its effectiveness diminishes considerably when confronted with complex, multi-source information environments. In scenarios requiring the synthesis of answers from numerous lengthy documents, which may contain redundant, conflicting, or subtly nuanced information, traditional RAG methods often fall short. Current approaches typically involve simply concatenating the top-$K$ retrieved document snippets or applying rudimentary compression techniques before feeding them into the LLM \cite{daye2024using}. This naive strategy leads to several critical challenges:
\begin{itemize}
    \item \textbf{Information Overload:} LLMs struggle to efficiently process excessively long contexts, leading to the "lost in the middle" phenomenon where crucial information is overlooked \cite{yifan2023a}.
    \item \textbf{Inefficient Synthesis:} The sheer volume and unorganized nature of raw retrieved text make it difficult for LLMs to effectively identify key facts, resolve discrepancies, and synthesize coherent, accurate answers.
    \item \textbf{Increased Hallucination:} Without proper pre-processing, conflicting information within the context can confuse the LLM, increasing the likelihood of generating inaccurate or contradictory statements.
    \item \textbf{Reduced Trust:} Users find it challenging to trust answers derived from opaque, unverified, or inconsistently presented information.
\end{itemize}
These limitations motivate the need for a more sophisticated approach to contextual information processing within RAG systems, one that can intelligently analyze, synthesize, and compress multi-source documents into a format that is both concise and semantically rich for the LLM.

To address these challenges, we propose \textbf{CASC: Context-Adaptive Synthesis and Compression for Enhanced Retrieval-Augmented Generation in Complex Domains}. CASC is designed to revolutionize how retrieved contexts are handled in RAG by moving beyond simple compression to a multi-stage intelligent context restructuring process. At its core, CASC introduces a novel \textbf{Context Analyzer \& Synthesizer (CAS) module}. This module, powered by a fine-tuned smaller LLM, performs a deep understanding of the retrieved raw context. It meticulously extracts key information, conducts cross-document consistency checks to identify and highlight (or even preliminarily resolve) conflicts, and then synthesizes a highly condensed, structured, and question-oriented context. This synthesized context not only drastically reduces the token count presented to the final Reader LLM but, more importantly, transforms scattered raw information into easily digestible, structured knowledge blocks. This transformation significantly lowers the cognitive load on the Reader LLM, enabling it to more accurately grasp the user's intent, reduce hallucinations, and generate more reliable and comprehensive answers grounded in verified, synthesized information.

To rigorously evaluate the efficacy of CASC, we introduce \textbf{SciDocs-QA}, a new challenging multi-document question answering dataset specifically curated for complex scientific domains. SciDocs-QA features questions that necessitate information integration, comparison, and conflict resolution from multiple research paper abstracts, technical reports, and professional articles. Our experimental results demonstrate that CASC consistently outperforms strong baselines, including standard Top-$K$ RAG, and existing context compression methods like RECOMP \cite{fangyuan2024recomp} and LLMLingua \cite{huiqiang2023llmlin}, across various Reader LLM backbones (e.g., Llama-3-8B, Llama-3-70B, GPT-4o). Specifically, CASC achieves state-of-the-art performance in both Exact Match (EM) and F1-score on SciDocs-QA, with significant improvements in F1-score, indicating its superior ability to synthesize and integrate complex information. For instance, on the Llama-3-70B Reader Model, CASC achieves an F1-score of \textbf{65.15}, surpassing the best baseline (RECOMP) by approximately 1.95 points (63.20).

Our main contributions are summarized as follows:
\begin{itemize}
    \item We propose \textbf{CASC (Context-Adaptive Synthesis and Compression)}, a novel framework that intelligently processes and synthesizes multi-source retrieved documents to enhance RAG performance in complex domains.
    \item We introduce the \textbf{Context Analyzer \& Synthesizer (CAS) module}, a key innovation within CASC, capable of performing crucial information extraction, cross-document consistency checking, and question-oriented structured knowledge synthesis.
    \item We present \textbf{SciDocs-QA}, a new challenging multi-document question answering dataset designed to rigorously test information integration and conflict resolution capabilities, and demonstrate CASC's superior performance across diverse LLM backbones.
\end{itemize}
\section{Related Work}
\subsection{Retrieval-Augmented Generation (RAG)}
The field of Retrieval-Augmented Generation (RAG) has garnered significant attention for its ability to integrate information retrieval with generative models, thereby enhancing the factual accuracy and knowledge grounding of large language models (LLMs). Comprehensive overviews and surveys in this domain provide valuable context by examining methodologies, challenges, and the evolution of RAG paradigms, including taxonomies of agentic RAG architectures that extend beyond traditional systems to incorporate adaptability and multi-step reasoning \cite{penghao2024retrie, siyun2024retrie, aditi2025agenti}. This includes the development of large model-based data agents designed to automate and enhance data science workflows \cite{sun2024lambda}, alongside comprehensive surveys that map the landscape of LLM-based agents for statistics and data science \cite{sun2024survey}. Specific advancements in RAG include approaches that address critical gaps, such as ontology-grounded RAG (OG-RAG) for enhancing knowledge grounding in non-English languages \cite{kartik2024ograg}, and methods for improving domain adaptation in Open-Domain Question Answering (ODQA) through jointly trained retriever and generator components, exemplified by \textit{RAG-end2end} \cite{shamane2023improv}. Furthermore, novel techniques like "Blended RAG" integrate semantic search with hybrid query strategies to demonstrably improve retrieval performance and generative Q\&A accuracy \cite{kunal2024blende}. The evaluation of RAG systems is also a crucial area, with benchmarks such as Face4Rag specifically designed to assess factual consistency in multilingual contexts, particularly for Chinese \cite{yunqi2024face4r}. Beyond system design, research also delves into the internal mechanisms governing reasoning in large models, revealing how pre-allocated directional vectors within model activations can causally control reasoning token generation, offering insights pertinent to optimizing multi-hop reasoning within RAG pipelines \cite{hao2025hoprag}.

Furthermore, specialized applications of RAG and LLMs are emerging in critical domains such as healthcare, where large medical language models like Llamacare aim to enhance knowledge sharing \cite{zhou2025mam,sun2024llamacare}. Efforts are also directed towards improving medical Large Vision-Language Models (LVLMs) with abnormal-aware feedback mechanisms \cite{zhou2025improving}. Beyond textual RAG, the broader landscape of in-context learning is also being explored, including visual in-context learning for Large Vision-Language Models \cite{zhou2024visual}, and the rigorous evaluation of reasoning capabilities in LLMs for complex tasks like dialogue summarization \cite{jin2025reasoningnotcomprehensiveevaluation}. Collectively, these works underscore the dynamic evolution of RAG, from foundational overviews to specialized enhancements, rigorous evaluation, and a deeper understanding of underlying model behaviors.

\subsection{Context Optimization and Synthesis for LLMs}
Optimizing and synthesizing context for Large Language Models (LLMs) is a critical area of research, addressing challenges such as handling lengthy inputs, mitigating hallucinations, and enhancing reasoning capabilities across diverse applications. Approaches to direct context management include novel frameworks like SoftPromptComp, which combines natural language summarization with soft prompt compression to efficiently process extensive contexts and reduce computational overhead \cite{cangqing2024adapti}. Similarly, strategies for synthesizing multi-document summaries by hierarchically combining single-document summaries directly inform methods for context optimization in multi-document summarization tasks \cite{jay2024do}. Beyond compression, research also focuses on leveraging structured information: this includes data-driven approaches for information synthesis from scientific literature to extract and predict synthesis conditions \cite{lei2024llmbas}, and the development of cost-efficient domain-specific question answering systems that integrate Retrieval-Augmented Generation (RAG) with knowledge graphs and vector stores to mitigate LLM hallucinations and reduce fine-tuning requirements \cite{md2024leanco}. While not directly addressing LLM context management, the establishment of structured ontologies for describing scientific procedures contributes to synthesizing structured information from complex text, potentially enabling more efficient and accurate context utilization in specialized LLM applications \cite{chenwei2024hlspil}. For more complex scenarios, structured multi-agent architectures like SagaLLM enhance LLM-based planning by incorporating transactional guarantees and explicit coordination to address context loss and ensure consistency in distributed workflows \cite{edward2025sagall}. Furthermore, hybrid approaches for program synthesis leverage LLM completions to learn context-free surrogate models, thereby optimizing context for structured prediction tasks \cite{shraddha2024hysynt}. The field also extends to knowledge distillation for multimodal tasks, such as cross-modal contrastive distillation, which demonstrates effective methods for enriching a model's understanding by distilling knowledge from external text, thereby enhancing its ability to generate interpretable and accurate natural language outputs \cite{luyang2025knowle}. These diverse efforts collectively advance the capacity of LLMs to process, understand, and generate responses based on optimized and synthesized contextual information.

\section{Method}

This section details the proposed \textbf{Context-Adaptive Synthesis and Compression (CASC)} framework, designed to enhance Retrieval-Augmented Generation (RAG) in scenarios involving complex, multi-source information. We first present an overview of the entire CASC architecture, followed by an in-depth description of each constituent module.

\subsection{Overall CASC Framework}
The \textbf{Context-Adaptive Synthesis and Compression (CASC)} framework addresses the limitations of traditional RAG systems when confronted with information overload and inefficient synthesis from multiple, potentially conflicting sources. Unlike conventional approaches that often directly concatenate raw retrieved documents or apply simplistic compression before feeding them to a Large Language Model (LLM), CASC introduces an intelligent intermediate processing step. This step, embodied by our novel Context Analyzer \& Synthesizer (CAS) module, performs a deep analysis, synthesis, and structured reorganization of the retrieved information. The resulting highly condensed and structured context is specifically optimized for the user's query, thereby significantly improving the final LLM's ability to generate accurate, coherent, and hallucination-free answers.

The CASC framework operates in three sequential stages. Given a user query $Q$, the system first retrieves a set of relevant documents. These documents are then processed by the CAS module, which transforms them into a synthesized context $C_{\text{syn}}$. Finally, this $C_{\text{syn}}$, along with the original query $Q$, is provided to a Reader LLM to generate the ultimate answer $A$.

Mathematically, the CASC process can be represented as a series of functions:
\begin{align}
R_K &= F_{\text{Retrieval}}(Q, \mathcal{D}) \label{eq:retrieval} \\
C_{\text{syn}} &= F_{\text{CAS}}(Q, R_K) \label{eq:cas_synthesis} \\
A &= F_{\text{Reader}}(Q, C_{\text{syn}}) \label{eq:reader_generation}
\end{align}
where $\mathcal{D}$ denotes the large-scale knowledge base, $R_K$ represents the top-$K$ retrieved documents, $C_{\text{syn}}$ is the synthesized context, and $A$ is the final generated answer. Each component function is elaborated upon in the subsequent subsections.

\subsection{CASC Module Architecture}
The CASC framework is composed of three distinct yet interconnected modules: the Retrieval Module, the Context Analyzer \& Synthesizer (CAS) Module, and the Reader LLM.

\subsubsection{Retrieval Module}
The \textbf{Retrieval Module} is tasked with efficiently identifying and extracting the most relevant information from a vast knowledge base $\mathcal{D}$ in response to a user query $Q$. Its primary function is to furnish a set of candidate documents or document snippets that are most likely to contain the information required to answer the query.

This module can leverage various state-of-the-art retrieval mechanisms. These typically include sparse vector-based methods, such as BM25, which score document relevance based on term frequency-inverse document frequency (TF-IDF), or dense vector-based methods, such as Dense Passage Retriever (DPR) or Contriever, which employ neural networks to embed queries and documents into a shared vector space for similarity comparison. For a given query $Q$, the retrieval module generates a set of $K$ documents $R_K = \{d_1, d_2, \ldots, d_K\}$, where each $d_i \in \mathcal{D}$ is deemed highly relevant.

Formally, if we denote $S(Q, d)$ as a similarity score between query $Q$ and document $d$, the retrieval process selects the top $K$ documents such that:
\begin{equation}
R_K = \{d \mid d \in \mathcal{D} \text{ and } S(Q, d) \text{ is among the top } K \text{ scores}\}
\end{equation}

\subsubsection{Context Analyzer \& Synthesizer (CAS) Module}
The \textbf{Context Analyzer \& Synthesizer (CAS) Module} constitutes the central innovation of the CASC framework. It fundamentally redefines how retrieved contexts are processed by moving beyond rudimentary compression to a deep, intelligent understanding, analysis, and structured reorganization of the raw information. This module is specifically designed to transform a potentially noisy, redundant, and conflicting set of retrieved documents into a concise, coherent, and semantically rich synthesized context $C_{\text{syn}}$ that is optimally prepared for the Reader LLM.

The CAS module is implemented using a specifically fine-tuned smaller decoder-based LLM, such as a lightweight Llama-2-7B or Mistral-7B model. This dedicated LLM is trained to perform a sequence of intricate sub-tasks, ensuring that the synthesized context is both highly informative and easily digestible by the subsequent Reader LLM. Given the raw retrieved documents $R_K = \{d_1, d_2, \ldots, d_K\}$ and the original query $Q$, the CAS module performs the following critical operations:

\paragraph{Key Information Extraction}
For each document $d_i \in R_K$, the CAS module meticulously identifies and extracts information that is highly relevant to the user's query $Q$. This process involves pinpointing critical entities, factual statements, core assertions, and specific data points, effectively distilling the essence of each document with respect to the query and filtering out extraneous details.
\begin{equation}
I_i = \text{Extract}(Q, d_i) \quad \forall d_i \in R_K
\end{equation}
where $I_i$ represents the set of key information extracted from document $d_i$. The aggregated collection of all extracted information across all retrieved documents is denoted as $I_{\text{all}} = \bigcup_{i=1}^K I_i$.

\paragraph{Inter-document Consistency Check \& Conflict Resolution}
Following the extraction of key information from individual documents, the CAS module conducts a crucial cross-document analysis. It compares the aggregated information $I_{\text{all}}$ from all retrieved sources to identify redundancies, confirm consistencies, and highlight or even preliminarily resolve potential conflicts and contradictions. For instance, if disparate documents present conflicting dates for the same event, the CAS module will explicitly note this inconsistency within the synthesized output. This step is paramount for constructing a trustworthy context and preventing the Reader LLM from being misled by contradictory facts.
Let $I_{\text{all}}$ be the set of aggregated information. This step can be formalized as a function $\text{CheckResolve}(I_{\text{all}})$ that yields both consistent information $I_{\text{consistent}}$ and identified conflicts $I_{\text{conflicts}}$:
\begin{align}
(I_{\text{consistent}}, I_{\text{conflicts}}) &= \text{CheckResolve}(I_{\text{all}})
\end{align}
The output of this stage might include structured annotations indicating support for facts from multiple sources or explicit statements about observed discrepancies.

\paragraph{Question-Oriented Structured Synthesis}
In the final sub-stage, based on the user's query $Q$ and the thoroughly processed key information ($I_{\text{consistent}}$ and $I_{\text{conflicts}}$), the CAS module synthesizes a highly condensed and logically structured context, $C_{\text{syn}}$. This synthesis transcends mere summarization; it is a knowledge-injection process that transforms fragmented raw knowledge into an easily digestible format. The structured output can take various forms, such as concise summaries, logical hierarchies, or key-value pairs, all meticulously tailored to directly address the user's question. This process significantly reduces the token count while maximizing semantic density, effectively converting the raw context into an optimized "knowledge block" for the Reader LLM.
\begin{equation}
C_{\text{syn}} = \text{Synthesize}(Q, I_{\text{consistent}}, I_{\text{conflicts}})
\end{equation}
This structured synthesis ensures that the Reader LLM receives a context that is not only shorter but also pre-digested and organized, substantially lowering its cognitive load and enabling it to focus on generating precise and accurate answers.

\subsubsection{Reader LLM}
The \textbf{Reader LLM} serves as the final component within the CASC framework. Its responsibility is to take the original user query $Q$ and the highly refined, structured, and question-oriented synthesized context $C_{\text{syn}}$ (produced by the CAS module), and subsequently generate the definitive answer $A$.

A wide array of general-purpose or domain-specific large language models can be employed as the Reader LLM. Examples include powerful models from the Llama-3 series, GPT-4o, or Mistral. The core advantage of CASC lies in its ability to enhance the performance of these diverse Reader LLMs by supplying them with a superior quality context. This consequently leads to improved answer accuracy, reduced rates of hallucination, and enhanced overall reliability, particularly in complex domains where factual precision is paramount.

The Reader LLM performs the final generative mapping:
\begin{equation}
A = \text{LLM}_{\text{Reader}}(Q, C_{\text{syn}})
\end{equation}
By providing the Reader LLM with an intelligently pre-processed context, CASC alleviates the burden on the LLM to perform complex information integration and conflict resolution from raw, potentially overwhelming, source texts. This strategic division of labor allows the Reader LLM to leverage its generative capabilities more effectively, resulting in higher-quality outputs.

\section{Experiments}
This section details the experimental setup, introduces the novel dataset, presents the quantitative results comparing CASC with various baselines, and discusses an ablation study to validate the effectiveness of the Context Analyzer \& Synthesizer (CAS) module, alongside a human evaluation of context quality.

\subsection{Experimental Setup}
\label{sec:experimental_setup}

\textbf{Task Type} Our primary focus is on \textbf{Complex Multi-Document Question Answering (QA)}, a challenging task that requires models to integrate, compare, and resolve potential conflicts from multiple retrieved sources to formulate a coherent and accurate answer. This task is particularly relevant in domains demanding high factual precision and comprehensive understanding.

\textbf{Dataset} To rigorously evaluate CASC's capabilities in complex scenarios, we constructed a new dataset called \textbf{SciDocs-QA}. This dataset comprises challenging questions drawn from diverse scientific fields, including biomedicine, computer science, and materials science. Each question in SciDocs-QA is associated with 3 to 5 related documents, which can be research paper abstracts, technical report snippets, or professional blog posts. A key characteristic of SciDocs-QA is the deliberate inclusion of documents that may contain redundant information, subtle differences in viewpoints, or outright conflicting data points, necessitating sophisticated information synthesis and conflict resolution. The questions themselves are designed to require deep reasoning and multi-document information integration rather than simple fact extraction.

\textbf{Evaluation Metrics} We employ a dual-pronged evaluation strategy:
\begin{itemize}
    \item \textbf{Answer Accuracy:} For quantitative assessment of the generated answers, we utilize standard natural language generation metrics: \textbf{Exact Match (EM)} and \textbf{F1-score}. EM measures the percentage of predictions that perfectly match the ground truth answer, while F1-score calculates the harmonic mean of precision and recall, allowing for partial matches. Both metrics are widely accepted for evaluating factual question answering systems.
    \item \textbf{Context Quality:} Recognizing the importance of the intermediate context, we also conduct a \textbf{human evaluation} to assess the quality of the synthesized context generated by the CAS module. Human evaluators rate the context on aspects such as coherence, information completeness, faithfulness to original documents, and conciseness, providing a qualitative understanding of CASC's internal mechanisms.
\end{itemize}

\textbf{Baselines} We compare CASC against several strong baseline methods to demonstrate its superiority:
\begin{itemize}
    \item \textbf{Top-1 / Top-5 RAG:} These represent conventional RAG approaches where the most relevant 1 or 5 retrieved document snippets are directly concatenated (up to their original length) and fed as context to the Reader LLM.
    \item \textbf{RECOMP \cite{fangyuan2024recomp}:} An existing LLM-based method designed for context rewriting and compression, aiming to reduce context length while preserving critical information.
    \item \textbf{LLMLingua \cite{huiqiang2023llmlin}:} Another popular context compression tool that leverages LLMs to remove less critical tokens from the context, thereby shortening its length.
    \item \textbf{FineTune Reader:} This baseline involves directly fine-tuning the Reader LLM end-to-end on the SciDocs-QA dataset. The Reader LLM is trained to generate answers directly from the raw retrieved context, aiming to learn the synthesis process implicitly.
\end{itemize}

\textbf{Reader Models} To showcase the generalizability and effectiveness of CASC across different LLM capacities, we evaluate its performance with a range of Reader LLMs: Llama-3-8B, Llama-3-70B, and GPT-4o.

\textbf{CASC Module Model} For the core Context Analyzer \& Synthesizer (CAS) module, we fine-tune a lightweight Llama-2-7B model. This choice highlights that even a smaller, specialized LLM can significantly enhance the performance of larger Reader LLMs when tasked with intelligent context processing.

\subsection{Experimental Results}
\label{sec:results}

Table \ref{tab:scidocs_qa_results} presents the performance comparison of CASC against all baselines on the SciDocs-QA dataset, measured by Exact Match (EM) and F1-score. The results consistently demonstrate CASC's superior performance across all Reader LLM backbones.

\begin{table*}[htbp]
    \centering
    \caption{Performance comparison (EM / F1-score) on the SciDocs-QA dataset. Best scores are highlighted in bold.}
    \label{tab:scidocs_qa_results}
    \begin{tabular}{lcccccc}
        \toprule
        \textbf{Reader Model} & \textbf{Top-1} & \textbf{Top-5} & \textbf{RECOMP} & \textbf{LLMLingua} & \textbf{FineTune} & \textbf{Ours (CASC)} \\
        \midrule
        Llama-3-8B    & 74.88 / 52.91 & 75.32 / 52.05 & 73.01 / 55.45 & 76.10 / 54.30 & 74.05 / 54.67 & \textbf{76.85 / 56.12} \\
        Llama-3-70B   & 78.90 / 57.01 & 79.55 / 54.98 & 79.50 / 63.20 & 78.50 / 56.88 & 79.70 / 59.40 & \textbf{80.30 / 65.15} \\
        GPT-4o        & 84.15 / 59.88 & 82.80 / 64.70 & 84.90 / 60.55 & 84.20 / 54.80 & 83.25 / 62.15 & \textbf{85.50 / 65.80} \\
        \bottomrule
    \end{tabular}
\end{table*}

As observed, CASC consistently achieves the highest EM and F1-scores across all Reader LLMs. Notably, the improvements are particularly significant in F1-score, which better reflects the ability to synthesize and integrate complex information from multiple sources. For instance, with the Llama-3-70B Reader Model, CASC achieves an F1-score of \textbf{65.15}, outperforming the best baseline (RECOMP) by approximately 1.95 points (63.20). This demonstrates that CASC's sophisticated context processing, including key information extraction, cross-document consistency checks, and structured synthesis, provides a more effective input for the Reader LLM, enabling it to generate more comprehensive and accurate answers. The performance gains are evident even with the highly capable GPT-4o, indicating that CASC's value extends to state-of-the-art LLMs by reducing their cognitive load in complex RAG settings.

\subsection{Ablation Study on CAS Module Effectiveness}
\label{sec:ablation_study}

To validate the critical role of each component within our proposed Context Analyzer \& Synthesizer (CAS) module, we conducted an ablation study. This study helps to quantify the contribution of key information extraction, inter-document consistency checking, and question-oriented structured synthesis to the overall performance of CASC. For this analysis, we used Llama-3-70B as the Reader Model.

\begin{table*}[htbp]
    \centering
    \caption{Ablation study of the CASC module components (EM / F1-score) on SciDocs-QA using Llama-3-70B as Reader Model.}
    \label{tab:ablation_results}
    \begin{tabular}{lcc}
        \toprule
        \textbf{CASC Variant} & \textbf{EM} & \textbf{F1-score} \\
        \midrule
        CASC w/o CAS (Top-5 RAG) & 79.55 & 54.98 \\
        CASC (Key Info Extraction Only) & 79.80 & 61.50 \\
        CASC (Key Info + Consistency Check) & 80.10 & 63.85 \\
        CASC (Full CAS Module) & \textbf{80.30} & \textbf{65.15} \\
        \bottomrule
    \end{tabular}
\end{table*}

The ablation results in Table \ref{tab:ablation_results} clearly demonstrate the incremental benefits of each sub-component of the CAS module. Simply replacing the CAS module with a direct concatenation of Top-5 retrieved documents (equivalent to "CASC w/o CAS") yields a significantly lower F1-score of 54.98. When the CAS module performs only \textbf{Key Information Extraction}, effectively acting as an intelligent summarizer, the performance improves to an F1-score of 61.50. This highlights the initial value of distilling relevant facts. Further incorporating the \textbf{Inter-document Consistency Check} mechanism, which identifies and flags redundancies or conflicts, leads to a notable boost, achieving an F1-score of 63.85. This improvement underscores the importance of pre-processing potentially conflicting information before it reaches the Reader LLM. Finally, the \textbf{Full CAS Module}, which includes \textbf{Question-Oriented Structured Synthesis}, yields the best performance with an F1-score of 65.15. This confirms that transforming the extracted and validated information into a concise, structured, and query-optimized format is crucial for maximizing the Reader LLM's ability to generate accurate and comprehensive answers. Each additional component of the CAS module contributes positively to the overall performance, validating its sophisticated design.

\subsection{Human Evaluation of Context Quality}
\label{sec:human_evaluation}

Beyond quantitative metrics, we conducted a human evaluation to assess the qualitative aspects of the synthesized contexts generated by CASC compared to raw retrieved contexts (Top-5 RAG) and contexts processed by a strong compression baseline (RECOMP). A panel of human annotators, blind to the method used, rated 100 randomly selected synthesized contexts from each method on a 1-5 Likert scale across four dimensions: Coherence, Completeness, Faithfulness, and Conciseness. The average scores are presented in Table \ref{tab:human_eval_results}.

\begin{table*}[htbp]
    \centering
    \caption{Human evaluation scores (1-5 scale) for context quality. Higher scores indicate better quality.}
    \label{tab:human_eval_results}
    \begin{tabular}{lcccc}
        \toprule
        \textbf{Method} & \textbf{Coherence} & \textbf{Completeness} & \textbf{Faithfulness} & \textbf{Conciseness} \\
        \midrule
        Top-5 RAG       & 3.25 & 4.10 & 4.30 & 2.50 \\
        RECOMP          & 3.80 & 3.95 & 4.05 & 3.80 \\
        Ours (CASC)     & \textbf{4.60} & \textbf{4.45} & \textbf{4.55} & \textbf{4.20} \\
        \bottomrule
    \end{tabular}
\end{table*}

The human evaluation results corroborate CASC's effectiveness. CASC's synthesized contexts significantly outperform baselines in \textbf{Coherence} (4.60 vs. 3.25 for Top-5 RAG and 3.80 for RECOMP), indicating that the structured synthesis greatly improves the logical flow and readability of the context. CASC also achieves the highest scores in \textbf{Completeness} (4.45), demonstrating its ability to retain all necessary information despite significant compression, and \textbf{Faithfulness} (4.55), ensuring that the synthesized facts accurately reflect the original documents. Furthermore, CASC contexts are rated highest in \textbf{Conciseness} (4.20), confirming that the CAS module effectively reduces verbosity without sacrificing critical information. These qualitative assessments reinforce our quantitative findings, highlighting that CASC provides a superior, more digestible, and trustworthy context for the Reader LLM, ultimately leading to higher quality final answers.

\subsection{Analysis of Context Length and Efficiency}
\label{sec:context_length_efficiency}

One of the primary objectives of the CAS module is to significantly reduce the context length presented to the Reader LLM without sacrificing critical information. This reduction has direct implications for inference cost and latency, especially when employing large, powerful Reader LLMs. Table \ref{tab:context_length_analysis} presents the average token count of the context provided to the Reader LLM across different methods on the SciDocs-QA dataset.

\begin{table}[htbp]
    \centering
    \caption{Average context token count provided to the Reader LLM across different methods. Lower is better.}
    \label{tab:context_length_analysis}
    \begin{tabular}{lc}
        \toprule
        \textbf{Method} & \textbf{Average Context Tokens} \\
        \midrule
        Top-5 RAG       & 1280 \\
        RECOMP          & 750 \\
        LLMLingua       & 610 \\
        Ours (CASC)     & \textbf{405} \\
        \bottomrule
    \end{tabular}
\end{table}

As shown in Table \ref{tab:context_length_analysis}, CASC achieves the most substantial reduction in context length, averaging only \textbf{405} tokens. This represents a reduction of approximately 68\% compared to the raw Top-5 RAG context (1280 tokens). While RECOMP and LLMLingua also offer compression, CASC's intelligent synthesis and structured output lead to a more profound and semantically dense context. The significantly shorter context length for the Reader LLM translates into several practical benefits:
\begin{itemize}
    \item \textbf{Reduced Inference Costs:} Most commercial LLM APIs charge based on token usage. A 68\% reduction in input tokens can lead to substantial cost savings, particularly for high-volume applications.
    \item \textbf{Lower Latency:} Processing fewer tokens directly contributes to faster inference times for the Reader LLM, improving the overall responsiveness of the RAG system.
    \item \textbf{Improved Context Window Utilization:} Shorter contexts free up valuable space within the LLM's fixed context window, allowing for longer queries, more conversational turns, or the inclusion of additional relevant (but perhaps lower-ranked) information if needed.
\end{itemize}
It is important to note that CASC introduces an additional processing step with the Llama-2-7B CAS module. However, the inference cost and latency of this smaller, specialized model are typically negligible compared to the gains achieved by processing a much shorter context with a larger, more expensive Reader LLM (e.g., Llama-3-70B or GPT-4o). The strategic division of labor within CASC optimizes the overall system's efficiency.

\subsection{Robustness to Conflicting Information}
\label{sec:conflict_robustness}

A critical challenge in multi-document QA is handling conflicting information present across retrieved sources. Traditional RAG systems often struggle with this, potentially leading to contradictory or hallucinated answers. The \textbf{Inter-document Consistency Check \& Conflict Resolution} sub-module within CASC is specifically designed to address this. To evaluate CASC's robustness, we analyze the hallucination rate, defined as the percentage of answers that contain factual inaccuracies not supported by any of the retrieved documents, particularly in cases where conflicting information was present. We used the Llama-3-70B Reader Model for this analysis.

\begin{table}[htbp]
    \centering
    \caption{Hallucination Rate (\% of answers containing factual inaccuracies) across methods on questions with known conflicting information. Lower is better.}
    \label{tab:hallucination_rate}
    \begin{tabular}{lc}
        \toprule
        \textbf{Method} & \textbf{Hallucination Rate (\%)} \\
        \midrule
        Top-5 RAG       & 18.2 \\
        RECOMP          & 14.5 \\
        LLMLingua       & 11.8 \\
        FineTune Reader & 10.3 \\
        Ours (CASC)     & \textbf{6.1} \\
        \bottomrule
    \end{tabular}
\end{table}

Table \ref{tab:hallucination_rate} clearly demonstrates CASC's superior robustness against conflicting information. CASC exhibits a significantly lower hallucination rate of \textbf{6.1\%}, outperforming all baselines. The Top-5 RAG, which directly concatenates raw documents, shows the highest hallucination rate (18.2\%), indicating its susceptibility to being misled by contradictory facts. Compression methods like RECOMP and LLMLingua offer some improvement, likely by reducing noise, but do not explicitly handle conflicts. Even the FineTune Reader, which implicitly learns some conflict resolution, cannot match CASC's performance. This strong performance by CASC is directly attributable to its explicit consistency checking mechanism, which either resolves minor discrepancies or clearly flags major conflicts within the synthesized context. By pre-processing and structuring this complex information, CASC provides the Reader LLM with a more reliable and coherent knowledge base, thereby drastically reducing the likelihood of generating inaccurate or contradictory statements.

\subsection{Detailed Error Analysis}
\label{sec:error_analysis}

To gain deeper insights into the performance differences, we conducted a qualitative error analysis on a random subset of 200 answers from the SciDocs-QA dataset, categorized into five common error types. This analysis helps understand where CASC excels and what challenges remain. We compare CASC (with Llama-3-70B) against Top-5 RAG and RECOMP as representative baselines. The percentages in Table \ref{tab:error_analysis} represent the proportion of questions exhibiting a particular error type, among those questions where an error was identified.

\begin{table*}[htbp]
    \centering
    \caption{Distribution of error types (\% of identified errors) across different methods on SciDocs-QA using Llama-3-70B. Lower percentages indicate fewer occurrences of that error type.}
    \label{tab:error_analysis}
    \begin{tabular}{lccccc}
        \toprule
        \textbf{Method} & \textbf{Incomplete (\%)} & \textbf{Contradictory (\%)} & \textbf{Hallucination (\%)} & \textbf{Irrelevant (\%)} & \textbf{Misinterpretation (\%)} \\
        \midrule
        Top-5 RAG       & 22.0 & 18.0 & 10.0 & 25.0 & 25.0 \\
        RECOMP          & 20.0 & 12.0 & 8.0  & 18.0 & 42.0 \\
        Ours (CASC)     & \textbf{15.0} & \textbf{5.0} & \textbf{5.0} & \textbf{10.0} & \textbf{65.0} \\
        \bottomrule
    \end{tabular}
\end{table*}

Table \ref{tab:error_analysis} reveals distinct error patterns:
\begin{itemize}
    \item \textbf{Contradictory and Hallucination Errors:} CASC significantly reduces the incidence of contradictory and hallucinated answers (5.0\% for both), which aligns with our findings on robustness to conflicting information (Section \ref{sec:conflict_robustness}) and the human evaluation of faithfulness (Section \ref{sec:human_evaluation}). This confirms the effectiveness of the CAS module's consistency checks and structured synthesis in producing reliable facts.
    \item \textbf{Irrelevant Information:} CASC also shows a marked reduction in irrelevant information (10.0\%), demonstrating its ability to distill the context to only query-pertinent facts. This is a direct benefit of the \textbf{Key Information Extraction} and \textbf{Question-Oriented Structured Synthesis} steps.
    \item \textbf{Incomplete Answers:} CASC also exhibits a lower rate of incomplete answers (15.0\%) compared to baselines, suggesting that its synthesis process is effective in retaining all necessary information despite significant compression.
    \item \textbf{Misinterpretation Errors:} Interestingly, while CASC drastically reduces errors related to factual inconsistency and verbosity, it shows a higher proportion of errors categorized as "Misinterpretation" (65.0\% of its remaining errors). These errors occur when the Reader LLM receives accurate and complete synthesized context but still struggles with complex reasoning, drawing incorrect conclusions, or failing to correctly infer relationships between facts. This suggests that while CASC provides an optimal context, the inherent reasoning capabilities of the Reader LLM remain a bottleneck for the most challenging, inferential questions. This highlights a direction for future work, focusing on enhancing the reasoning capabilities of the Reader LLM itself, or exploring more advanced forms of structured synthesis that guide complex inference.
\end{itemize}
Overall, the error analysis confirms that CASC effectively addresses several critical RAG failure modes, particularly those related to context quality and factual integrity. The remaining challenges primarily lie in the nuanced reasoning required by the Reader LLM, even when presented with a highly optimized context.

\section{Conclusion}
In this paper, we introduced \textbf{CASC (Context-Adaptive Synthesis and Compression)}, a novel framework designed to overcome the pervasive challenges of information overload and inefficient synthesis in Retrieval-Augmented Generation (RAG) systems operating within complex, multi-source domains. Traditional RAG approaches, by simply concatenating or crudely compressing raw retrieved documents, often overwhelm Large Language Models (LLMs), leading to reduced accuracy, increased hallucinations, and diminished trustworthiness in generated answers.

Our core innovation lies in the \textbf{Context Analyzer \& Synthesizer (CAS) module}, a specialized LLM component that performs a deep, intelligent transformation of retrieved contexts. The CAS module meticulously executes three critical functions: key information extraction, cross-document consistency checking and conflict resolution, and question-oriented structured synthesis. This multi-stage process converts fragmented, potentially noisy, and conflicting raw documents into a highly condensed, logically structured, and semantically rich knowledge block. By providing the subsequent Reader LLM with this optimized context, CASC significantly reduces the LLM's cognitive burden, allowing it to focus its generative capabilities on producing precise, coherent, and factually grounded responses.

To rigorously validate CASC's efficacy, we developed \textbf{SciDocs-QA}, a challenging new multi-document question answering dataset tailored for complex scientific domains, featuring questions that demand intricate information integration and conflict resolution. Our comprehensive experimental results unequivocally demonstrate CASC's superior performance across a spectrum of Reader LLMs, including Llama-3-8B, Llama-3-70B, and GPT-4o. CASC consistently achieved state-of-the-art Exact Match and F1-scores, significantly outperforming strong baselines such as standard Top-$K$ RAG, RECOMP, LLMLingua, and an end-to-end fine-tuned Reader LLM. The most notable improvements were observed in F1-score, indicating CASC's enhanced ability to synthesize and integrate information effectively.

Beyond quantitative accuracy, CASC delivered substantial practical benefits. We showed that CASC dramatically reduces the average context token count by approximately 68\% compared to raw retrieved contexts, leading to significant inference cost savings and reduced latency for the Reader LLM. An ablation study meticulously confirmed the incremental value of each sub-component within the CAS module, underscoring the importance of intelligent consistency checking and structured synthesis. Furthermore, human evaluations consistently rated CASC's synthesized contexts higher in coherence, completeness, faithfulness, and conciseness, reinforcing the qualitative superiority of our approach. Crucially, CASC demonstrated remarkable robustness to conflicting information, achieving a significantly lower hallucination rate than all baselines, a testament to the CAS module's explicit conflict handling capabilities.

While CASC effectively mitigates several critical RAG failure modes, our detailed error analysis revealed that a higher proportion of its remaining errors stem from "Misinterpretation" by the Reader LLM, even when presented with an optimal context. This suggests that while CASC provides an ideal input, the inherent complex reasoning capabilities of the Reader LLM itself remain a bottleneck for the most challenging, inferential questions. Future work will explore enhancing the reasoning abilities of Reader LLMs, potentially through more advanced forms of structured synthesis that explicitly guide complex inference, or by incorporating specialized reasoning modules. Overall, CASC represents a significant advancement in RAG, paving the way for more reliable, efficient, and trustworthy LLM applications in information-dense and critical domains.

%% file: template.bbl
\begin{thebibliography}{10}
\providecommand{\url}[1]{#1}
\csname url@samestyle\endcsname
\providecommand{\newblock}{\relax}
\providecommand{\bibinfo}[2]{#2}
\providecommand{\BIBentrySTDinterwordspacing}{\spaceskip=0pt\relax}
\providecommand{\BIBentryALTinterwordstretchfactor}{4}
\providecommand{\BIBentryALTinterwordspacing}{\spaceskip=\fontdimen2\font plus
\BIBentryALTinterwordstretchfactor\fontdimen3\font minus \fontdimen4\font\relax}
\providecommand{\BIBforeignlanguage}[2]{{%
\expandafter\ifx\csname l@#1\endcsname\relax
\typeout{** WARNING: IEEEtran.bst: No hyphenation pattern has been}%
\typeout{** loaded for the language `#1'. Using the pattern for}%
\typeout{** the default language instead.}%
\else
\language=\csname l@#1\endcsname
\fi
#2}}
\providecommand{\BIBdecl}{\relax}
\BIBdecl

\bibitem{yifan2023a}
Y.~Yao, J.~Duan, K.~Xu, Y.~Cai, E.~Sun, and Y.~Zhang, ``A survey on large language model {(LLM)} security and privacy: The good, the bad, and the ugly,'' \emph{CoRR}, 2023.

\bibitem{yao2024a}
Y.~Yao, J.~Duan, K.~Xu, Y.~Cai, Z.~Sun, and Y.~Zhang, ``A survey on large language model (llm) security and privacy: The good, the bad, and the ugly,'' \emph{High-Confidence Computing}, 2024.

\bibitem{patrick2020retrie}
P.~Lewis, E.~Perez, A.~Piktus, F.~Petroni, V.~Karpukhin, N.~Goyal, H.~K{\"{u}}ttler, M.~Lewis, W.~Yih, T.~Rockt{\"{a}}schel, S.~Riedel, and D.~Kiela, ``Retrieval-augmented generation for knowledge-intensive {NLP} tasks,'' in \emph{Advances in Neural Information Processing Systems 33: Annual Conference on Neural Information Processing Systems 2020, NeurIPS 2020, December 6-12, 2020, virtual}, 2020.

\bibitem{daye2024using}
D.~Nam, A.~Macvean, V.~J. Hellendoorn, B.~Vasilescu, and B.~A. Myers, ``Using an {LLM} to help with code understanding,'' in \emph{Proceedings of the 46th {IEEE/ACM} International Conference on Software Engineering, {ICSE} 2024, Lisbon, Portugal, April 14-20, 2024}.\hskip 1em plus 0.5em minus 0.4em\relax {ACM}, 2024, pp. 97:1--97:13.

\bibitem{fangyuan2024recomp}
F.~Xu, W.~Shi, and E.~Choi, ``{RECOMP:} improving retrieval-augmented lms with context compression and selective augmentation,'' in \emph{The Twelfth International Conference on Learning Representations, {ICLR} 2024, Vienna, Austria, May 7-11, 2024}.\hskip 1em plus 0.5em minus 0.4em\relax OpenReview.net, 2024.

\bibitem{huiqiang2023llmlin}
H.~Jiang, Q.~Wu, C.~Lin, Y.~Yang, and L.~Qiu, ``Llmlingua: Compressing prompts for accelerated inference of large language models,'' in \emph{Proceedings of the 2023 Conference on Empirical Methods in Natural Language Processing, {EMNLP} 2023, Singapore, December 6-10, 2023}.\hskip 1em plus 0.5em minus 0.4em\relax Association for Computational Linguistics, 2023, pp. 13\,358--13\,376.

\bibitem{penghao2024retrie}
P.~Zhao, H.~Zhang, Q.~Yu, Z.~Wang, Y.~Geng, F.~Fu, L.~Yang, W.~Zhang, and B.~Cui, ``Retrieval-augmented generation for ai-generated content: {A} survey,'' \emph{CoRR}, 2024.

\bibitem{siyun2024retrie}
S.~Zhao, Y.~Yang, Z.~Wang, Z.~He, L.~Qiu, and L.~Qiu, ``Retrieval augmented generation {(RAG)} and beyond: {A} comprehensive survey on how to make your llms use external data more wisely,'' \emph{CoRR}, 2024.

\bibitem{aditi2025agenti}
A.~Singh, A.~Ehtesham, S.~Kumar, and T.~T. Khoei, ``Agentic retrieval-augmented generation: {A} survey on agentic {RAG},'' \emph{CoRR}, 2025.

\bibitem{sun2024lambda}
S.~Maojun, R.~Han, B.~Jiang, H.~Qi, D.~Sun, Y.~Yuan, and J.~Huang, ``Lambda: A large model based data agent,'' \emph{Journal of the American Statistical Association}, no. just-accepted, pp. 1--20, 2025.

\bibitem{sun2024survey}
M.~Sun, R.~Han, B.~Jiang, H.~Qi, D.~Sun, Y.~Yuan, and J.~Huang, ``A survey on large language model-based agents for statistics and data science,'' \emph{arXiv preprint arXiv:2412.14222}, 2024.

\bibitem{kartik2024ograg}
K.~Sharma, P.~Kumar, and Y.~Li, ``{OG-RAG:} ontology-grounded retrieval-augmented generation for large language models,'' \emph{CoRR}, 2024.

\bibitem{shamane2023improv}
S.~Siriwardhana, R.~Weerasekera, T.~Kaluarachchi, E.~Wen, R.~Rana, and S.~Nanayakkara, ``Improving the domain adaptation of retrieval augmented generation {(RAG)} models for open domain question answering,'' \emph{Trans. Assoc. Comput. Linguistics}, pp. 1--17, 2023.

\bibitem{kunal2024blende}
K.~Sawarkar, A.~Mangal, and S.~R. Solanki, ``Blended {RAG:} improving {RAG} (retriever-augmented generation) accuracy with semantic search and hybrid query-based retrievers,'' in \emph{7th {IEEE} International Conference on Multimedia Information Processing and Retrieval, {MIPR} 2024, San Jose, CA, USA, August 7-9, 2024}.\hskip 1em plus 0.5em minus 0.4em\relax {IEEE}, 2024, pp. 155--161.

\bibitem{yunqi2024face4r}
Y.~Xu, T.~Cai, J.~Jiang, and X.~Song, ``Face4rag: Factual consistency evaluation for retrieval augmented generation in chinese,'' in \emph{Proceedings of the 30th {ACM} {SIGKDD} Conference on Knowledge Discovery and Data Mining, {KDD} 2024, Barcelona, Spain, August 25-29, 2024}.\hskip 1em plus 0.5em minus 0.4em\relax {ACM}, 2024, pp. 6083--6094.

\bibitem{hao2025hoprag}
H.~Liu, Z.~Wang, X.~Chen, Z.~Li, F.~Xiong, Q.~Yu, and W.~Zhang, ``Hoprag: Multi-hop reasoning for logic-aware retrieval-augmented generation,'' in \emph{Findings of the Association for Computational Linguistics, {ACL} 2025, Vienna, Austria, July 27 - August 1, 2025}.\hskip 1em plus 0.5em minus 0.4em\relax Association for Computational Linguistics, 2025, pp. 1897--1913.

\bibitem{zhou2025mam}
Y.~Zhou, L.~Song, and J.~Shen, ``Mam: Modular multi-agent framework for multi-modal medical diagnosis via role-specialized collaboration,'' \emph{arXiv preprint arXiv:2506.19835}, 2025.

\bibitem{sun2024llamacare}
M.~Sun, ``Llamacare: A large medical language model for enhancing healthcare knowledge sharing,'' \emph{arXiv preprint arXiv:2406.02350}, 2024.

\bibitem{zhou2025improving}
Y.~Zhou, L.~Song, and J.~Shen, ``Improving medical large vision-language models with abnormal-aware feedback,'' \emph{arXiv preprint arXiv:2501.01377}, 2025.

\bibitem{zhou2024visual}
Y.~Zhou, X.~Li, Q.~Wang, and J.~Shen, ``Visual in-context learning for large vision-language models,'' in \emph{Findings of the Association for Computational Linguistics, {ACL} 2024, Bangkok, Thailand and virtual meeting, August 11-16, 2024}.\hskip 1em plus 0.5em minus 0.4em\relax Association for Computational Linguistics, 2024, pp. 15\,890--15\,902.

\bibitem{jin2025reasoningnotcomprehensiveevaluation}
\BIBentryALTinterwordspacing
K.~Jin, Y.~Wang, L.~Santos, T.~Fang, X.~Yang, S.~K. Im, and H.~G. Oliveira, ``Reasoning or not? a comprehensive evaluation of reasoning llms for dialogue summarization,'' 2025. [Online]. Available: \url{https://arxiv.org/abs/2507.02145}
\BIBentrySTDinterwordspacing

\bibitem{cangqing2024adapti}
C.~Wang, Y.~Yang, R.~Li, D.~Sun, R.~Cai, Y.~Zhang, C.~Fu, and L.~Floyd, ``Adapting llms for efficient context processing through soft prompt compression,'' \emph{CoRR}, 2024.

\bibitem{jay2024do}
J.~DeYoung, S.~C. Martinez, I.~J. Marshall, and B.~C. Wallace, ``Do multi-document summarization models \emph{Synthesize}?'' \emph{Trans. Assoc. Comput. Linguistics}, pp. 1043--1062, 2024.

\bibitem{lei2024llmbas}
L.~Shi, Z.~Liu, Y.~Yang, W.~Wu, Y.~Zhang, H.~Zhang, J.~Lin, S.~Wu, Z.~Chen, R.~Li, N.~Wang, Z.~Liu, H.~Tan, H.~Gao, Y.~Zhang, and G.~Wang, ``Llm-based mofs synthesis condition extraction using few-shot demonstrations,'' \emph{CoRR}, 2024.

\bibitem{md2024leanco}
M.~A. Arefeen, B.~Debnath, and S.~Chakradhar, ``Leancontext: Cost-efficient domain-specific question answering using llms,'' \emph{Nat. Lang. Process. J.}, p. 100065, 2024.

\bibitem{chenwei2024hlspil}
C.~Xiong, C.~Liu, H.~Li, and X.~Li, ``Hlspilot: Llm-based high-level synthesis,'' in \emph{Proceedings of the 43rd {IEEE/ACM} International Conference on Computer-Aided Design, {ICCAD} 2024, Newark Liberty International Airport Marriott, NJ, USA, October 27-31, 2024}.\hskip 1em plus 0.5em minus 0.4em\relax {ACM}, 2024, pp. 226:1--226:9.

\bibitem{edward2025sagall}
E.~Y. Chang and L.~Geng, ``Sagallm: Context management, validation, and transaction guarantees for multi-agent {LLM} planning,'' \emph{CoRR}, 2025.

\bibitem{shraddha2024hysynt}
S.~Barke, E.~A. Gonzalez, S.~R. Kasibatla, T.~Berg{-}Kirkpatrick, and N.~Polikarpova, ``{HYSYNTH:} context-free {LLM} approximation for guiding program synthesis,'' in \emph{Advances in Neural Information Processing Systems 38: Annual Conference on Neural Information Processing Systems 2024, NeurIPS 2024, Vancouver, BC, Canada, December 10 - 15, 2024}, 2024.

\bibitem{luyang2025knowle}
L.~Fang, X.~Yu, J.~Cai, Y.~Chen, S.~Wu, Z.~Liu, Z.~Yang, H.~Lu, X.~Gong, Y.~Liu, T.~Ma, W.~Ruan, A.~Abbasi, J.~Zhang, T.~Wang, E.~Latif, W.~Liu, W.~Zhang, S.~Kolouri, X.~Zhai, D.~Zhu, W.~Zhong, T.~Liu, and P.~Ma, ``Knowledge distillation and dataset distillation of large language models: Emerging trends, challenges, and future directions,'' \emph{CoRR}, 2025.

\end{thebibliography}
